\documentclass[journal]{IEEEtran}
\usepackage{bm}
\usepackage{amsmath}
\usepackage{amssymb}
\usepackage{amsthm}
\usepackage{mathrsfs}
\usepackage{enumerate}
\usepackage{multirow}
\usepackage{color}
\usepackage{threeparttable}
\usepackage{subfigure}

\usepackage[square, comma, sort&compress, numbers]{natbib}
\usepackage[pagebackref=false,breaklinks=true,letterpaper=true,colorlinks,bookmarks=false]{hyperref}

\usepackage{times}
\usepackage{breakurl}
\usepackage{array}
\usepackage{verbatim}
\usepackage{algorithm}
\usepackage{bbding}
\usepackage{algpseudocode}
\usepackage{lettrine}

\ifCLASSINFOpdf
  \usepackage[pdftex]{graphicx}
  \usepackage{graphics}
  \usepackage{graphicx}
  \usepackage{epsfig}
  \usepackage{epstopdf}
  \graphicspath{{./fig/}}
  \DeclareGraphicsExtensions{.pdf,.jpeg,.png,.eps}
\else
  \usepackage[dvips]{graphicx}
  \graphicspath{{./fig/}}
  \DeclareGraphicsExtensions{.eps}
\fi

\ifCLASSINFOpdf
\else
\fi
\begin{document}
%
\title{HSGNet: Object Re-identification with Hierarchical Similarity Graph Network}
%
%
%

\author{Fei Shen,
        Mengwan Wei,
        and Junchi Ren

}

%
%

\markboth{Journal of \LaTeX\ Class Files,~Vol.~14, No.~8, August~2015}%
{Shell \MakeLowercase{\textit{et al.}}: Bare Demo of IEEEtran.cls for IEEE Journals}
%



\maketitle

\begin{abstract}
Object re-identification method is made up of backbone network, feature aggregation, and loss function.
However, most backbone networks lack a special mechanism to handle rich scale variations and mine discriminative feature representations.
In this paper, we firstly design a hierarchical similarity graph module (HSGM) to reduce the conflict of backbone and re-identification networks.
The designed HSGM  builds a rich hierarchical graph to mine the mapping  relationships between global-local and local-local.
Secondly, we divide the feature map along with the spatial and channel directions in each hierarchical graph.
The HSGM applies the spatial features and channel features extracted from different locations as nodes, respectively, and utilizes the similarity scores between nodes to construct spatial and channel similarity graphs.
During the learning process of HSGM, we utilize a learnable parameter to re-optimize the importance of each position, as well as evaluate the correlation between different nodes.
Thirdly, we develop a novel hierarchical similarity graph network (HSGNet) by embedding the HSGM in the backbone network.
Furthermore, HSGM can be easily embedded into backbone networks of any depth to improve object re-identification ability.
Finally, extensive experiments on three large-scale object datasets demonstrate that the proposed HSGNet is superior to state-of-the-art object re-identification approaches.
\end{abstract}

\begin{IEEEkeywords}
Hierarchical similarity graph, object re-identification, deep
learning.
\end{IEEEkeywords}

%
\IEEEpeerreviewmaketitle

\section{Introduction}
\label{sec:intro}

Object re-identification \cite{zhang2022unbiased, huang2021hierarchical, zheng2021viewpoint, liu2021prgcn, liu2022modeling, hpgn, strcmap, spindle, emrn, li2021diverse, he2019foreground} focuses on retrieval objects of interest from massive traffic data, playing a great potential for intelligent transportation systems.
With the development of classification tasks, the object re-identification community has become more attractive.
Existing object re-identification networks usually utilize backbone networks in classification networks to extract object features.
However, these backbone networks were initially designed for classification tasks and can not deal well with problems that are similar between intra-classes.
Therefore, mining discriminative features are crucial in improving object re-identification performance.

\begin{figure}[tp]
    \centering
    \includegraphics[width=0.9\linewidth]{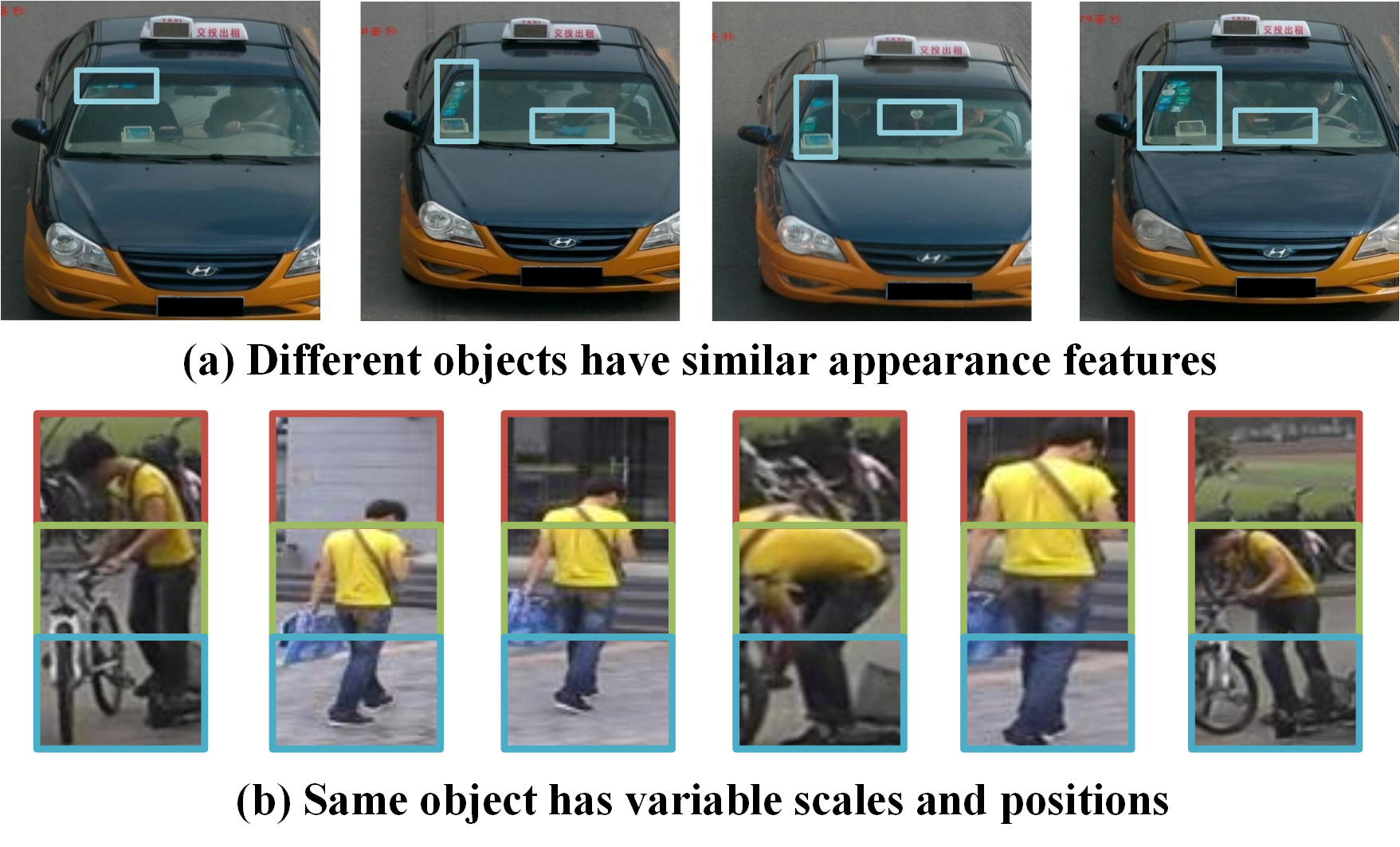}
    \caption{ {\color{black}
    The illustration of two critical  problem for object re-identification.}}
\label{fig:demo}
\end{figure}

Deep learning based object re-identification models \cite{labnet, hsgm, git} consist of backbone networks, feature aggregation architectures, and loss function.
In backbone networks, lots of mature deep learning networks, such as ResNeSt \cite{resnest}, SeNet \cite{senet}, and Res2Net \cite{res2net} can be adopted in object re-identification. However, as shown in Fig. \ref{fig:demo} (a),
since object being captured by different cameras
installed at different city locations, the images of different objects have very highly similar appearance features on the object's re-identification task;
On the contrary, in some local areas, such as the annual inspection sign of the vehicle's front window, the lights, the hairstyle of pedestrians, and the shoe area have obvious differences.
Therefore, how to mine the saliency of the object to emphasize the discriminative local area is of great significance for object re-identification.

In the re-identification research field, the straightforward approach \cite{par, strcmap, pamtri, spindle, git, pcb, hpm, vpm} is to divide vehicles or pedestrians into several regions/keypoints, and then study features independently on each regions/keypoints.
Depending on whether additional annotations are required, we can subdivide the method into partition-based and partition-free methods.
The partition-based methods denote partition with the help of additional information, e.g., multi-DNN fusion siamese neural network (MFSNN) \cite{mfsnn} part-aligned representations (PAR) \cite{par}, local feature preservation (LFP) \cite{strcmap},  partial attention and multi-attribute learning network (PAMLNet) \cite{tumrani2020partial} and SpindleNet \cite{spindle}.
The partition-free methods represent tutoring without additional information, e.g., contrastive attention module network (CAMNet) \cite{camnet}, part-based convolutional baseline (PCB) \cite{pcb}, efficient multi-resolution network (emrn) \cite{emrn} , hybrid pyramidal graph network (HPGN) \cite{hpgn}, visibility-aware part model (VPM) \cite{vpm}.
However, both types of methods have obvious drawbacks for object re-identification.
The partition-based methods \cite{par, strcmap, pamtri, spindle} face the problem of additional manual labeling and requiring a lot of computing resources.
As shown in Fig. \ref{fig:demo} (b), although partition-free methods \cite{pcb, hpm, vpm} do not require extra manual annotation, their performance is vulnerable to pose and scale changes.  Therefore, there are large intra-class
variations in pedestrian and vehicle images, which cause object re-identification to be very challenging.


For that, this paper designs a hierarchical similarity graph module (HSGM) to reduce the conflict of backbone and classfication networks.
The designed HSGM first builds a rich hierarchical graph to mine the mapping relationships between global-local and local-local.
The HSGM applies the spatial features and channel features extracted from different locations as nodes, respectively, and utilizes the similarity scores between nodes to construct spatial and channel similarity graphs.
During the learning process of HSGM, we utilize a learnable parameter to re-optimize the importance of each position, as well as evaluate the correlation between different nodes.
In addition, we develop a novel hierarchical similarity graph network (HSGNet) by embedding the HSGM in the backbone network.
Furthermore, HSGM can be easily embedded into backbone networks of any depth to improve object re-identification ability.

The HSGNet can effectively mine object features without additional human-assisted annotation.
The main reasons are summarized as follows:
(i) For highly similar appearance features of the different classes, the HSGNet can construct a node to represent a local region (e.g., lights or logos) in a similarity graph (i.e., spatial and channel). Then, it can learn a node’s feature representation from the node (e.g., lights or logos) and the node’s neighbors (e.g., roof and door).
Since we consider both internal and external relationships between nodes, this enables each local node feature to be enhanced.
In other words, we do not treat local features in isolation, but consider the mixed features of the local features and its surrounding nodes.
(ii) For different pose and position changes of the same object, the HSGNet can dynamically optimize the most representative areas (i.e., local and global) under this background of the hierarchical graph, which focuses on the discriminative features.
In other words, we need to pay more attention to the saliency or representative features of the object through this relationships of global-local and local-local.

{\color{black}
The main contributions of this paper can be summarized as follows:
\vspace{-.1cm}
\begin{itemize}
\item   A plug-and-play hierarchical similarity graph module (HSGM) is proposed to reduce the conflict of backbone and classification networks and aggregate the node's features via spatial and channel directions.

\item  According to the proposed HSGM, we develop a novel hierarchical similarity graph network (HSGNet) to improve the performance for object re-identification.
This is the first work that mines the important features via similarity graphs between the nodes to the best of our knowledge.


\item
We have conducted extensive experiments and achieved promising performance gains on three large-scale object datasets.
Then, a large number of ablation studies also verify the effectiveness of the core mechanisms in the HSGNet for object re-identification.

\end{itemize}
}




\section{Related Work}\label{sec:rw}
Object re-identification progress is reviewed from two aspects: (1)feature extraction and (2) loss functions.
\subsection{Feature Extraction}
Due to the development of upstream backbone networks, such as ResNet \cite{resnet}, SeNet \cite{senet}  and Res2Net \cite{res2net}, the re-identification community has grown very rapidly. Unfortunately, the re-identification task belongs to a kind of fine-grained image retrieval and inter-classes objects have highly similar appearance features.
 Moreover, the backbone networks cannot satisfactorily address the difference in local features since these methods are developed for classification tasks.
 For that, based on these classical backbone networks, \cite{strcmap, hpgn, pcb} have developed many partition methods for object re-identification. Among them, we can subdivide the partition method into partition-based and partition-free methods.
 Ye et al. \cite{ye2020dynamic} used the attention
mechanism on two-stream network to learn intra-modal difference
and cross-modal relationship to improve feature alignment
The LFP \cite{strcmap}  employs a fast R-CNN detector to narrow down the potential search area, and then uses the re-identification model to extract global features.
Based on the two-stream network, Zhang et al. \cite{zhang2021global} separated the feature to three
part and six part respectively, and attempted to align global and local features at the same time.
The adaptive attention vehicle re-identification (AAVER) \cite{aaver} employs key-point detection module to localizing the local features and use adaptive key-point selection module to learning the relationship of parts.

The similar approach is also used
in  partition-free methods, which learn local features to achieve more detailed
feature alignment.
For example,
the HPGN \cite{hpgn} proposes a  pyramidal graph network (PGN) to explore the spatial meaning of different regional features for vehicle re-identification.
The PCB \cite{pcb} utilizes a unified block strategy and uses independent convolution modules to extract features from each block for person re-identification.
The quadruple directional deep learning networks (QD-DLF) \cite{qddlf} proposes four novelty directional pooling layers to resist the negative impact of viewpoint changes on vehicle re-identification by fusing features from different orientations.

\subsection{Hierarchical Similarity Graph Module}
\begin{figure}[tp]
    \centering
    \includegraphics[width=0.9\linewidth]{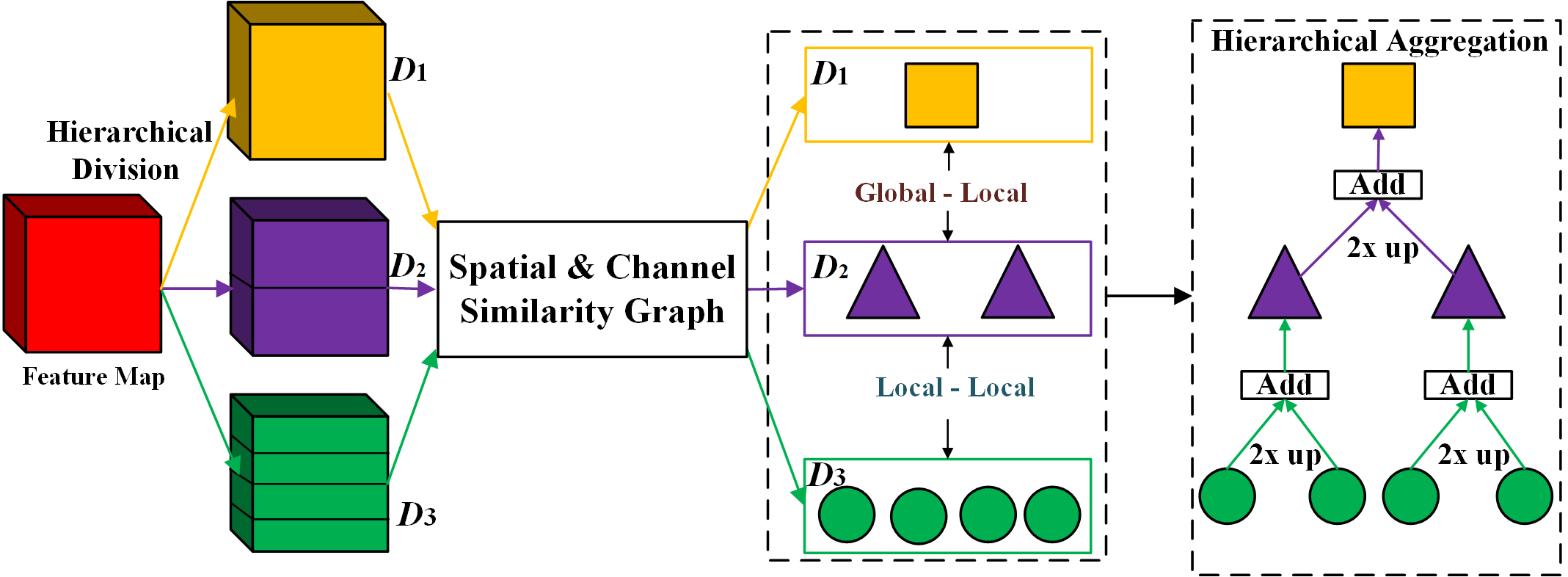}
    \caption{ {\color{black}
    The diagrams of hierarchical similarity graph module (HSGM).
}}
\label{fig:hg}
\end{figure}

\begin{figure*}[tp]
    \centering
    \includegraphics[width=0.95\linewidth]{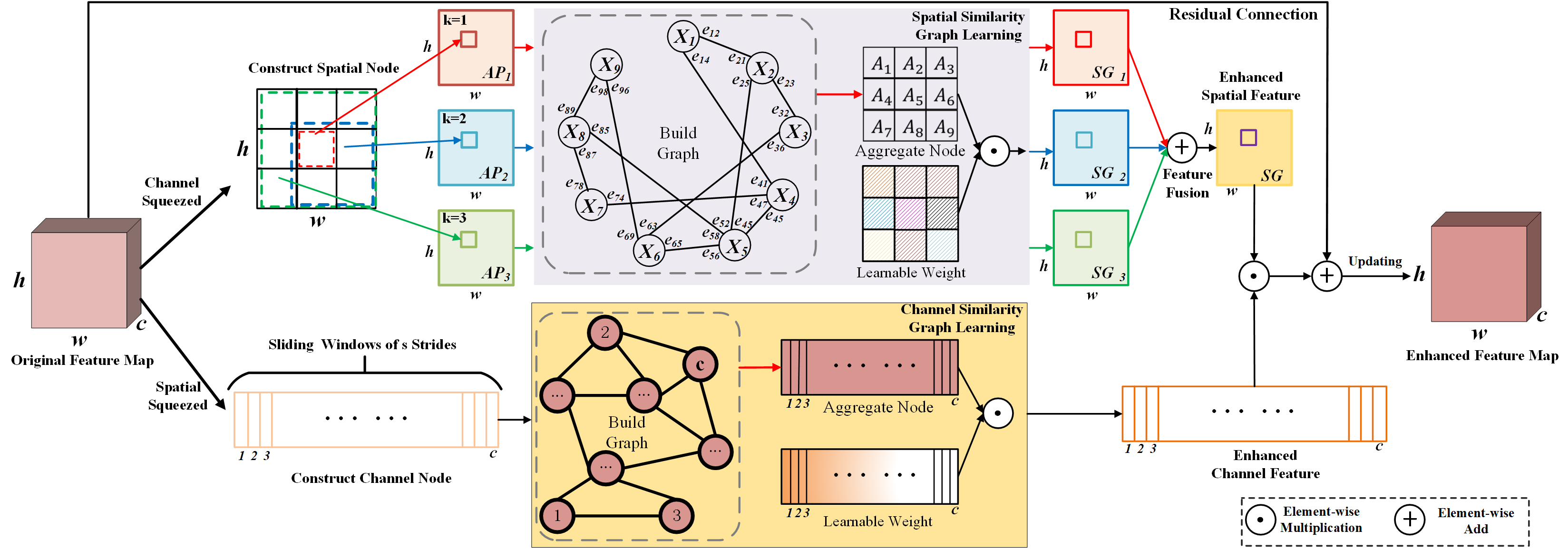}
    \caption{ The proposed of the similarity graph (SG) learning process, consists of the spatial similarity graph and channel similarity graph learning. The similarity graph learning contains constructed node, build graph, aggregate node, enhanced feature, and residual connection. }
\label{fig:mcsg}
\end{figure*}

\subsection{Loss Functions}
Loss functions for object re-identification can be divided into three types.
(1) The identification loss function based on cross entropy loss \cite{ram}, which aims to maximize the posterior probability of each object subject. The EMRN \cite{emrn} proposes a multi-resolution features dimension uniform module and multi cross entropy losses to fix dimensional features from images of varying resolutions.
(2) The classification loss function, such as contrastive loss \cite{contrasloss, camnet} function and hybrid similarity function \cite{hybrid}, which is with the goal of letting similarities of positive pairs (i.e., object subjects of the same class label) are positive while making similarities of negative pairs (i.e., object subjects of different class labels) are negative. For example, the contrastive attention module network (CAMNet) proposes two different mode loss to assess one part feature’s importance based on all parts.
(3) The ranking loss function \cite{facenet,triplet},  which focuses on pulling distances of positive pairs as close as possible and pushing distances of negative pairs as far as possible.
Recently object re-identification methods prefer to jointly use cross entropy loss \cite{deepid} and triplet loss \cite{facenet,triplet} or its variants \cite{triemd,drdl,ggl,gste,mgr,tvt,tamr,c2f,vami,mrl,vanet}. Since the cross entropy loss is beneficial to ensures a high convergence speed, while the triplet loss pulls distances of positive pairs as close as possible and pushes distances of negative pairs as far as possible.

\section{Methodology}\label{sec:method}

As shown in Fig. \ref{fig:hg}, the hierarchical similarity graph module (HSGM) consists of hierarchically dividing, spatial $\&$ channel similarity graph learning, and hierarchically aggregation. The core idea is that we do not treat local features in isolation, but consider the mixed features of the local features and its surrounding nodes. The processing of the proposed HSGM is as follows.
\subsubsection{Hierarchically Divide and Hierarchically Aggregation}
Suppose there is a feature map, we first copy it three times and hierarchically divide it into $D_{1}$, $D_{2}$, and $D_{3}$,
where $D_{1}$, $D_{2}$, and $D_{3}$ respectively have 1, 2, and 4 uniform feature maps.
It is worth noting that these uniformly divided feature maps (i.e., $D_{1}$, $D_{2}$, and $D_{3}$) have different local features or global features.
$D_{1}$ can represent the global features of the image. $D_{2}$ and $D_{3}$ can denote critical local features of the object.
We construct $D_{2}$ and $D_{1}$, $D_{2}$ and $D_{3}$ into global-local and local-local pairs, respectively, to focus on the different importance features of the object.
In other words, we need to pay more attention to the saliency or representative features of the object through this relationships of global-local and local-local.
Then, seven feature maps are independently fed to the similarity graphs, which contain a spatial similarity graph (SSG) learning branch and a channel similarity graph (CSG) learning branch, to obtain more discriminative features.
More detail about the SSG and CSG are described on Section \ref{ssg} and Section \ref{csg}.
Furthermore, the hierarchical aggregation up-sample $D_{3}$  by two times and adds $D_{2}$, and the same on $D_{2}$  is up-sampled two times and adds $D_{1}$.
Based on $D_{1}$, $D_{2}$, and $D_{3}$, the different feature maps are analogously fed to SSG and CSG independently, as shown in Figure \ref{fig:mcsg}.

\subsubsection{Spatial Similarity Graph Learning}\label{ssg}
To simplify the description, assume that $X \in R^{h\times w\times c}$ is a three-dimensional feature map in one of the different feature maps, where $h$, $w$, and $c$ represent the height, width, and channel number of $X$. The specific operation steps of the spatial similarity graphs learning branch are as follows.
We first perform channel squeezed on the feature map $X$, and the calculation process of channel squeezed is formulated as follows:
\begin{equation}\label{eq:cs}
S_{i, j}=\frac{1}{c} \sum_{m}^{c} X_{i, j, m},
\end{equation}
 where $X_{i, j, m}$ denotes represents the feature map value at row $i$, column $j$ on the $m$ channel, $S_{i, j}$ represents the feature map value of the corresponding spatial position after channel squeezed. Hence, we can obtain $S=\left\{S_{i, j} \mid i=1,2,3, \ldots, h, j=1,2,3, \ldots, w\right\} \in R^{h \times w}$.
Then, according to Eq. \eqref{eq:ap}, we apply three different average pooling (\emph{AP}) strategies to feature map $S$.
	\begin{equation}\label{eq:ap}
	S'_{k} = AP^{k}(S),
	\end{equation}
where $AP^{k}$ represents AP layer with $k \times k$ pixels, which set $k = 1, 2, 3$ in all experiments. Besides, we set $S' = \{S'_{1}, S'_{2}, S'_{3}\}$ to increase the perception domain at different areas.

According to the above $S'$, spatial similarity graphs $SSG_{k} = \{V^{s}_{k}, E^{s}_{k}\}$ are constructed to mine significant features.
The vector $V^{s}_{k}$ and  $E^{s}_{k}$ respectively denotes spatial node sets and spatial edge sets between different nodes.
The SSG (i.e., $SSG_{1}$, $SSG_{2}$ and $SSG_{3}$) of single-scale are fused into $SSG_{k}$ of multi-scale through element-wise addition operation, as shown in  Figure \ref{fig:mcsg}.

To better explain the principle, the subscript of $SSG_{k}$ and $X'_{k}$ are removed.
We can reshape  $X'_{k}$ as $X'\in R^{d}$ with $d=h \times w$ features.
Then, we contract each local feature as a node in HSGM.
Therefore, we can regard an instance SSG of local feature as $SSG = \{V^{s} \in R^{d} , E^{s} \in R^{d\times d }\}$, where $V^{s} = X'$.
Then, according to Eq. \eqref{eq:edge}, the SSG's edges are built.
	\begin{equation}\label{eq:edge}
	E^{s}_{i,j}=\left\{
	\begin{array}{cc}
	\frac{Exp({V^{s}_{i}})}{\sum_{1}^{\|N_{V^{s}_{i}}\|} Exp({V^{s}_{j}})}, & V^{s}_{j} \in N_{V^{s}_{i}}, \\
	0, & {otherwise},
	\end{array}\right.
	\end{equation}
where $i, j \in [1,2,...,d]$; $E^{s}_{i,j}$ is the edge between the node $V^{s}_{i}$ on the spatial feature map;
The $N_{V^{s}_{i}}$  and $\|N_{V^{s}_{i}}\|$  respectively denote the neighbor node set of $V^{s}_{i}$ and the number of nodes in $N_{V^{s}_{i}}$.
Specifically, we use the similarity between nodes to represent the value of their edges.
Then, the node $V^{s}$ aggregation is summarized as Eq. \eqref{eq:agg}.
	\begin{equation}\label{eq:agg}
	U^{s}=I^{s}V^{s}+E^{s}V^{s}=\left(I^{s}+E^{s}\right) V^{s}.
	\end{equation}
	It is worth noted that $I^{s}=diag(1,1,1,...,1) \in R^{d\times d}$ is an identity matrix.
	Then, each spatial aggregated node $A^{s}$ is further learned by applying Eq. \eqref{eq:dot}, as follows:
	\begin{equation}\label{eq:dot}
	A^{s}=U^{s}\odot {\varTheta}, 
	\end{equation}
where $\varTheta \in R^{d}$ and $\odot$ respectively represent a learnable parameter matrix and an element-wise multiplication operation.
From Eq. \eqref{eq:agg} and Eq. \eqref{eq:dot}, We can find that both the node itself and its neighbor nodes can be aggregated to provide more importance features to explore the saliency of the corresponding region.

Therefore,  we can convert $X'_{k}$ to $A^{s}_{k}$ in Figure \ref{fig:mcsg}.
 If there is no special instruction, we set $k =1, 2, 3$.
The SSG of multi-scale $A^{ms}$ is computed as:
	\begin{equation}\label{eq:sum}
	A^{ms}= \sum_{1}^{k}A^{s}_{k}.
	\end{equation}
Furthermore, we rearrange the spatial aggregated node $A^{ms}$ into a matrix of $1 \times h \times w$ according to memory continuity.

\begin{figure}[tp]
    \centering
    \includegraphics[width=0.9\linewidth]{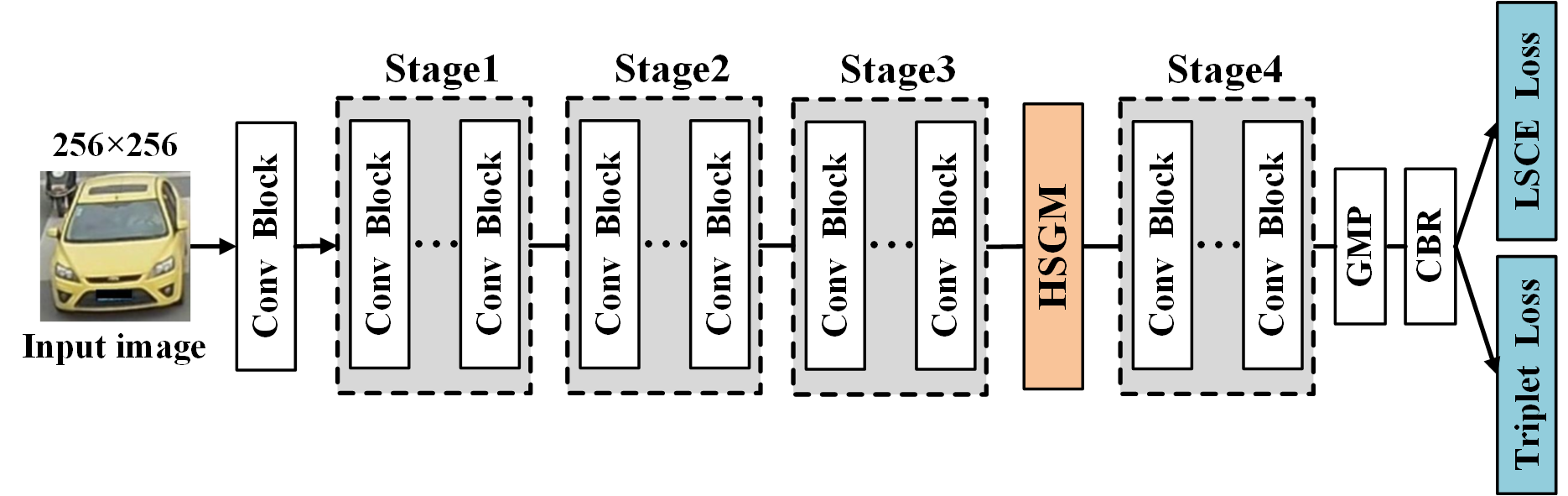}
    \caption{ {\color{black}
    The framework of the proposed hierarchical similarity graph network (HSGNet). }}
\label{fig:pipline}
\end{figure}

\subsubsection{Channel Similarity Graph Learning}\label{csg}
As shown in Fig. \ref{fig:mcsg}, we first perform spatial squeezed on the feature map $X$, and the calculation process of spatial squeezed is formulated as follows:
\begin{equation}\label{eq:cp}
C_{m}=\frac{1}{h \times w} \sum_{i=1}^{h} \sum_{j=1}^{w} X_{i, j, m}
\end{equation}
According to Eq. \eqref{eq:cp}, we can obtain $C=\left\{C_{m} \mid m=1,2,3, \ldots, c \right\} \in R^{c}$.
Then, we construct a channel similarity graph (CSG), $CSG = \{V^{c} \in R^{d} , E^{c} \in R^{c\times c }\}$,
where channel node sets  $V^{c}\left\{C_{m} \mid m=1,2,3, \ldots, c \right\}$ and the determination rule of channel edge sets $E^{c}$ is the same as Eq. \eqref{eq:edge}. It is determined by the sliding window of $s$ strides, and the neighbor nodes of the node at the boundary are correspondingly reduced.
The difference between $E^{c}$ and $E^{s}$ is that the former performs neighborhood determination in the channel dimension, while the latter conducts neighborhood determination in the spatial dimension.
Then, based on the CSG, the features of each channel node are aggregated to obtain the $U^{c}=\left(I^{c}+E^{c}\right) V^{c}$ aggregated by the channel nodes,
where $I^{c}$ is the identity matrix.
Further, each channel aggregated node $A^{c}$ is weighted with a learnable weight parameter $\delta \in R^{c}$, as shown in Eq. \eqref{eq:co}, to mine the importance in the channel dimension.
\begin{equation}\label{eq:co}
A^{c}=U^{c} \odot \delta
\end{equation}
The output $A^{ms}$ of the SSG and the output $A^{c}$ of the CSG are multiplied to obtain an enhanced feature map $A^{msc}$ , as Eq. \eqref{eq:multiplied}:
\begin{equation}\label{eq:multiplied}
A^{msc}=A^{ms} A^{c}
\end{equation}
Besides, to increase the SG's convergence in the training phase,  we enhance SG according to Eq. \eqref{eq:nonlinear}.
	\begin{equation}\label{eq:nonlinear}
	O=LeakyReLU \left(BN\left(A^{msc}\right)\right) ,
	\end{equation}
	where the $BN$ and $LeakyReLU$  respectively represent  the batch normalization and the leaky rectified linear unit.

 Lastly, inspired by residual mechanism \cite{resnet}, the  $X^{D}$ is formulated as:
	\begin{equation}\label{eq:res}
	X^{D}= O + X.
	\end{equation}

\subsection{Combined with Backbone Network}
As shown in Fig. \ref{fig:pipline}, the hierarchical similarity graph module (HSGM) can be easily embedded into existing backbone network at any depth.
Inspired by IANet \cite{ianet}, HSGM is placed at the down-sampling place of every residual group, which is the bottleneck of the block.
Based on the proposed HSGM, a flexible hierarchical similarity graph network (HSGNet) is proposed for object re-identification.
Specifically,
For HSGNet based object re-identification, we adopt a ResNet50 as the backbone network.
As shown in the Section \ref{sec:exp} for detailed analysis and comparison, our proposed HSGM is only inserted into the backbone network between stage3 and stage4 layers.
Following the strong baseline model  \cite{luo2019strong}, the 'last stride=1' training trick is used to set the last stride of the stage4 down-sampling to 1 for retaining more spatial information in the learned feature maps.
Besides, the HSGNet also contains a global max pooling (GMP) and a CBR composite unit.
The CBR denotes a convolutional layer using $1\times1$ sized filters, a batch normalization layer \cite{bnorm} and a ReLU \cite{relu} activation function.

Therefore, as shown in Fig. \ref{fig:pipline}, there are one triplet loss $L_{triplet}$ and one label smoothing cross entropy (LSCE) loss $L_{lsce}$ reconfigured to supervise the proposed HSGNet, thus the total loss function is formulated as follows:

	\begin{equation}\label{eq:totoal}
	{L_{total}} = \alpha L_{triplet} + \beta L_{lsce},
	\end{equation}
where $\alpha$ and $\beta$ are both a hype-parameter, used to control the balance between different losses.
Their default value are respectively set to 0.5 and 1.0 via cross-validation.
 The LSCE loss function $L_{lsce}$ is formulated as follows:
	\begin{equation}\label{eq:llsrs}
	{
		{L_{lsce}}(X,l) = \frac{{ - 1}}{N}\sum\limits_{i = 1}^N {\sum\limits_{j = 1}^K {\delta ({l_i},j)log(\frac{{{e^{W_j^{\rm{T}}{X_i}}}}}{{\sum\nolimits_{k = 1}^K {{e^{W_k^{\rm{T}}{X_i}}}} }})} },
	}
	\end{equation}
	\begin{equation}\label{eq:delta}
	{
		\delta ({l_i},j) =\left\{
		\begin{array}{cc}
		1 - \gamma  + \frac{\gamma }{K}, & j = {l_i}, \\
		\frac{\gamma }{K}, & {otherwise},
		\end{array}\right.
	}
	\end{equation}
	where $X$ and $l$ are respectively a training set and the class label information;
The $N$ is the number of training samples; $K$ is the number of classes; The $W=[W_1, W_2, W_3, ..., W_K]$ is a learnable parameter matrix.
The $(X_i, l_i)$ is the $i$-th training sample and $l_i \in \{1, 2, 3, ..., K\}$.
	
	The triplet loss function $L_{triplet}$  is formulated as follows:
	\begin{equation}\label{eq:triplet}
	{
		\begin{aligned}
		&{L_{triplet}}({X^a},{X^n},{X^p}) \\=
		&\frac{{ - 1}}{N}\sum\limits_{i = 1}^N {\max ({{\left\|{X_i^a - X_i^n} \right\|}_2}
		 -{{\left\|  {X_i^a - X_i^p} \right\|}_2} - \delta, 0)},
		\end{aligned}
	}
	\end{equation}
where $M$ and $({X^a},{X^n},{X^p})$ are respectively a set of training triplets and the number of training triplets.
The $\left\|\cdot\right\|_2$ denotes the Euclidean distance.
 For the $i$-th training triplet, $(X^a_i, X^p_i)$ and $(X^a_i, X^n_i)$ are a positive pair of the same class label and a negative pair of different class labels, respectively.
 The margin constant of triplet loss $L_{triplet}$ (i.e., $\delta$ of Eq. \ref{eq:triplet}) and the smoothing regularization degree of label smoothing cross entropy loss $L_{lsce}$ (i.e., $\gamma$ of Eq. \ref{eq:delta}) are respectively set to $1.2$ and $0.1$ in all experiments.

\section{Experiments and Analysis}\label{sec:exp}

{\color{black}

In this section, we evaluate our HSGNet on three well-known object re-identification datasets, namely, Market-1501 \cite{market1501}, VeRi776 \cite{veri776}, and VehicleID \cite{drdl}. For performance metrics, commonly-used mean average precision (mAP) \cite{veri776, drdl, market1501}, and cumulative match characteristic (CMC) \cite{veri776, drdl, market1501} curve are applied.

}

\subsection{Datasets}

{ \color{black}\textbf{\emph{Market-1501}} \cite{market1501} is a pedestrian Re-ID dataset captured by six
cameras. It contains 32,668 pedestrian images of 1,501 identities.
The training subset includes 12,936 images of 751 identities, while the test subset holds images of the rest 750 identities, i.e., 19,732 gallery pedestrian images and 3,368 query pedestrian images.
}

\begin{table}[tp]
	\centering
	\caption{
	The performance comparison of the proposed
method and state-of-the-art approaches on Market-1501 dataset.}\label{tab:market1501}
	\setlength{\tabcolsep}{5pt}
	\footnotesize
	\begin{tabular}{l||cccccc}
		\hline
  {Methods} &Rank1   &Rank5  & mAP        \\
	\hline
	SpindleNet \cite{spindle} &76.9 &91.5 &-   \\
    PAR \cite{par} &81.0 &92.0 &63.4   \\
    LSRO \cite{lsro} &84.0 &- &66.1   \\
    MultiScale \cite{multiscale} &88.9 &- &73.1 \\

    HA-CNN \cite{hacnn}  & 91.2&- &75.7    \\
    DuATM \cite{aaver} &91.4 &-  &76.6 \\
    PGFA \cite{pgfa} &91.2 &-  &76.8  \\
    VPM \cite{vpm} &93.0 &-  &80.8  \\
  	PCB \cite{pcb} &93.8 &97.5 &{81.6}  \\
    AANet \cite{aanet} &93.8 &-  &82.4  \\
    HPM \cite{hpm} &94.2 &97.5  &82.7 \\
    HOReID\cite{horeid} &94.2 &- &84.9  \\
    {SNR \cite{snr}}   & 94.4 & - &{84.7}   \\
    IANet \cite{ianet} &94.4 &-  &83.1  \\
    P$^2$-Net \cite{ppnet} &\textbf{95.2}& 98.2& 85.6 \\

        \hline
    BoT (Baseline) \cite{luo2019strong} &93.8 &97.3 &84.5  \\
    \textbf{HSGNet}   &{95.0} & \textbf{98.9} & \textbf{86.6} \\
    \hline
	\end{tabular}
\end{table}

{ \color{black}\textbf{\emph{VeRi776}}~\cite{veri776}  is a vehicle Re-ID dataset captured from real traffic scenarios by using 20 surveillance cameras. It totally consists of 776 subjects. The training subset contains 37,746 images of 576
subjects, and the rest 200 subjects are applied for the testing subset.
Furthermore, the test subset includes a probe subset of 1,678 images and a gallery subset of 11,579 images.
 }
\begin{table*}[tp]
	\centering
	\caption{The performance comparison of the proposed method and state-of-the-art approaches  on VeRi776 and VehicleID datasets.}\label{tab:veri776}
	\setlength{\tabcolsep}{5pt}
	\begin{tabular}{l||ccc|cccccc|cc}
		\hline
  \multirow{3}{*}{Methods} & \multicolumn{3}{c|} {\multirow{2}{*}{VeRi776}} & \multicolumn{8}{c}{VehicleID} \\
  &&&& \multicolumn{2}{c}{Test800} & \multicolumn{2}{c}{Test1600} &\multicolumn{2}{c}{Test2400}&\multicolumn{2}{c}{Average}\\ \cline{2-12}
  &Rank1   &Rank5  & mAP   &Rank1  & mAP   &Rank1  & mAP &Rank1  & mAP  &Rank1  & mAP  \\
	\hline
	DenseNet121 \cite{dense}  &80.27 &91.12&45.06 &66.10&68.85&67.39&69.45&63.07&65.37&65.52&67.89 \\ %
MGL \cite{mgl} & 86.10 &96.20 &65.00 &79.6 &82.1 &76.2 &79.6 &73.0 &75.5&76.26&79.06 \\
VAMI \cite{vami}          &85.92 &91.84 &61.32 &63.12 &- &52.87 &- &47.34 &-   &54.44&-\\
QD-DLF \cite{qddlf} &88.50&94.46&61.83 &72.32 &76.54 &70.66 &74.63 &64.14 &68.41&69.04&73.19\\
			RAM \cite{ram}         &88.60  & 94.00 &61.50 &75.20 &- &72.30 &- &67.70 &- &71.73&-\\
			ResNet101-AAVER \cite{aaver} &88.97 &94.70 &61.18 &74.69 &- &68.62 &- &63.54 &- &68.95&-\\
			Triplet Embedding \cite{triemd}    &90.23 &96.42 &67.55 &78.80&86.19&73.41&81.69&69.33&78.16&73.84&82.01 \\
			MRM \cite{mrm} &91.77 &95.82 &68.55 &76.64 &80.02 &74.20 &77.32 &70.86 &74.02 &73.90&77.12\\
			Part Regularization \cite{partreg} &94.30&98.70 &74.30  &78.40 &- &75.00 &- &74.20 &- &75.86&- \\
            EMRN \cite{emrn} &94.75&97.86&79.78   & 77.12 &80.70 &73.73 &77.00&70.71&74.14&73.85&77.28\\
			Appearance+License \cite{app} &95.41&97.38 &78.08   &79.50 &82.70 &76.90 &79.90 &74.80 &77.70&77.06&80.1\\

                \hline
            BoT (Baseline) \cite{luo2019strong} &95.23 &97.16 &77.31 &79.38 &82.47 &76.55 &79.61 &74.38 &77.26&77.02&79.78 \\
            \textbf{HSGNet}   &\textbf{96.94} & \textbf{98.73} & \textbf{80.65} &\textbf{82.77}&\textbf {85.34}&\textbf{78.81}&\textbf{83.28}&\textbf{76.65}&\textbf{80.63} &\textbf{79.41}&\textbf{82.97} \\

    \hline
	\end{tabular}
\end{table*}

\begin{table*}[tp]
	\centering
	\caption{
The comparison of the effect of HSGM on VeRi776 and Market-1501 datasets.}\label{tab:ablation}
	\begin{tabular}{c||ccc|ccc}
		\hline
		\multirow{2}{*}{
			\begin{tabular}{c}
			Methods
			\end{tabular}
		}
		&\multicolumn{3}{c|}{\multirow{1}*{VeRi776}}
		&\multicolumn{3}{c}{\multirow{1}*{Market-1501}}\\

		&Rank1     & Rank5 &mAP
        & Rank1 &Rank5     & mAP
	\\
\hline
	\multirow{1}*Baseline &95.23 & 97.16  &77.31 &93.8&97.3 &84.5\\
\hline
	\multirow{1}*{HSGM w/o Spatial and Channel Similarity Graph } &96.06 & 97.89  &79.38 &94.3&97.9 &85.2\\
\hline
	\multirow{1}*{HSGM w/o Hierarchical} &96.47 & 98.21  &80.03 &94.4&98.2 &85.8\\
\hline
	\multirow{1}*{HSGM w/o Res. Connection } &96.55 & 98.19  &79.95 &94.5&98.4 &86.1\\
\hline
	\multirow{1}*{HSGM w/o Spatial Similarity Graph} &96.67 & 98.43  &80.27 &94.6& 98.7 &86.3\\
\hline
	\multirow{1}*{HSGM w/o Channel Similarity Graph} &96.82 & 98.57  &80.44 &94.8& 98.9 &86.4\\
\hline
	\multirow{1}*{HSGNet} &\textbf{96.94} & \textbf{98.73}  &\textbf{80.65} &\textbf{95.0}& \textbf{98.9} &\textbf{86.6}\\
\hline
	\end{tabular}
\end{table*}

\textbf{\emph{VehicleID}} \cite{drdl}  is constructed from video sequences captured by
cameras in different locations, backgrounds and lighting
conditions, such as highway intersections, road crossings,
parking lots, etc, which totally includes $221,763$ images of $26,267$ classes. It has three test subsets, each of which randomly selects one image of a tested object to form the probe subset, and all remaining images form the gallery subset.

\subsection{Implementation Detail}

The operating system used in this experiment is Ubuntu16.04 and the deep learning toolbox is Pytorch with two V100 GPUs on the same machine for fair comparison.
Training configurations are summarized as follows.
(1) Common object re-identification training tricks are used, such as using ImageNet pretrained weights, random erasing, random flipping, and z-score normalization.
(2) We respectively resize the image resolution uniformly to 256 $\times $256 and 256 $\times $128 on vehicle (i.e., VeRi776 and VehicleID) and person (i.e., Market-1501) datasets.
(3) The mini-batch stochastic gradient descent (SGD) method is applied to train parameters.
The weight decays are set to be 5$\times$10$^{-4}$, and the momentums are set to be 0.9.
There are 120 epochs for the training process. The learning rates are initialized to 0.001, and linearly warmed up  to 0.01 in the first 10 epochs.
After warming up, the learning rates are maintained at 0.01 from
11st to 30th epochs. Then, the learning rates decays by 10\% every
20 epochs.
(4) Each mini-batch includes 32 subjects and each subject holds 4 images.
In the testing phase, the 2048-dimensional feature output by the GMP layer in Fig. \ref{fig:pipline} is used as the final image feature, and the cosine distance is used as the similarity measure for object re-identification.


\subsection{Comparison with State-of-the-art Methods}
\textbf{Comparisons on Market-1501.}
As shown in Table \ref{tab:market1501}, the proposed  the proposed hierarchical similarity graph network (HSGNet)  obtains the highest 95.0\% Rank1, 98.9\% Rank5 and 86.6\% mAP.
For example, compared with two partition-based methods, i,e, SpindleNet \cite{spindle} and PAR \cite{par}, the HSGNet is higher than 17.9\%  and 13.8\% on Rank1, respectively.
Considering the partition-free methods, i.e., AANet \cite{aanet} and IANet \cite{ianet}, it is still defeated 4.2\% and 3.5\% on mAP, respectively.
Although our method is slightly lower on Rank1 than P$^2$-Net \cite{ppnet}, which uses an additional semantic segmentation model, we defeat it by 1.0\% on mAP. Therefore, integrating Rank and mAP retrieval comparison results into account, our method remains competitive on Market-1501.

\textbf{Comparisons on VeRi776.}
The performance of HSGNet and multiple state-of-art methods are compared shown in Table \ref{tab:veri776}.
Firstly, Compared with those SOTA methods, the proposed HSGNet
obtains the best Rank1 (i.e., 96.94\%) and the highest mAP (i.e., 80.65\%).
Secondly, Compared with multi-scale methods, the HSGNet defeats the EMRN \cite{emrn} by 2.19\% in terms of Rank1 and 0.87\% in terms of mAP.
Thirdly, compared with partition-based methods (i.e., MRM \cite{mrm} and part regularization \cite{partreg}), the proposed HSGNet still defeat 5.17\% and 2.64 \% on Rank1, respectively.
Lastly, the HSGNet still respectively outperforms the method of additional vehicle attribute annotations, i.e., Appearance+License \cite{app} by 2.57\% and 1.53\% larger on mAP and Rank1 identification rate. It is worth noting that the proposed HSGNet method does not require license plates. Therefore, our proposed HSGNet
is superior than these state-of-the-art approaches on VeRi776 dataset in both partition-based and partition-free methods.

\begin{figure}[tp]
	\centering
	\includegraphics[width=.9\linewidth]{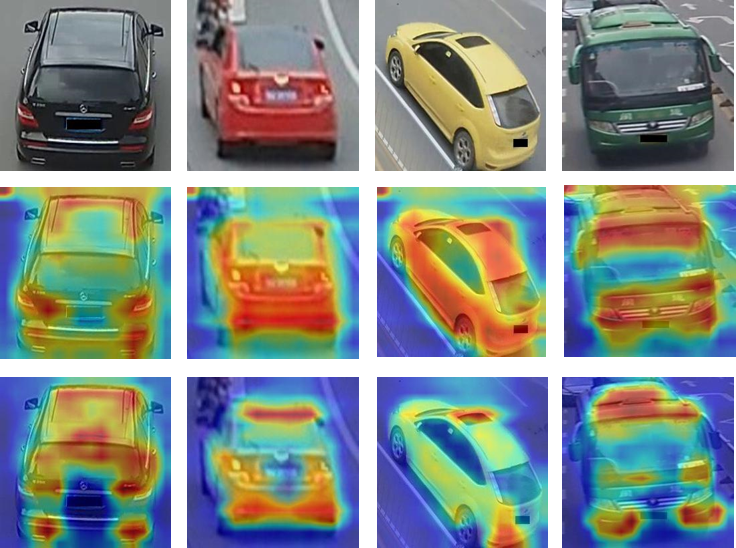}
	\caption{The grad class activation maps (Grad-CAM) visualization of feature maps on VeRi776 dataset. The first, second, and third rows show the original image, Baseline, and HSGNet, respectively.}
	\label{fig:vis}
\end{figure}

\begin{figure}[tp]
    \centering
    \includegraphics[width=.9\linewidth]{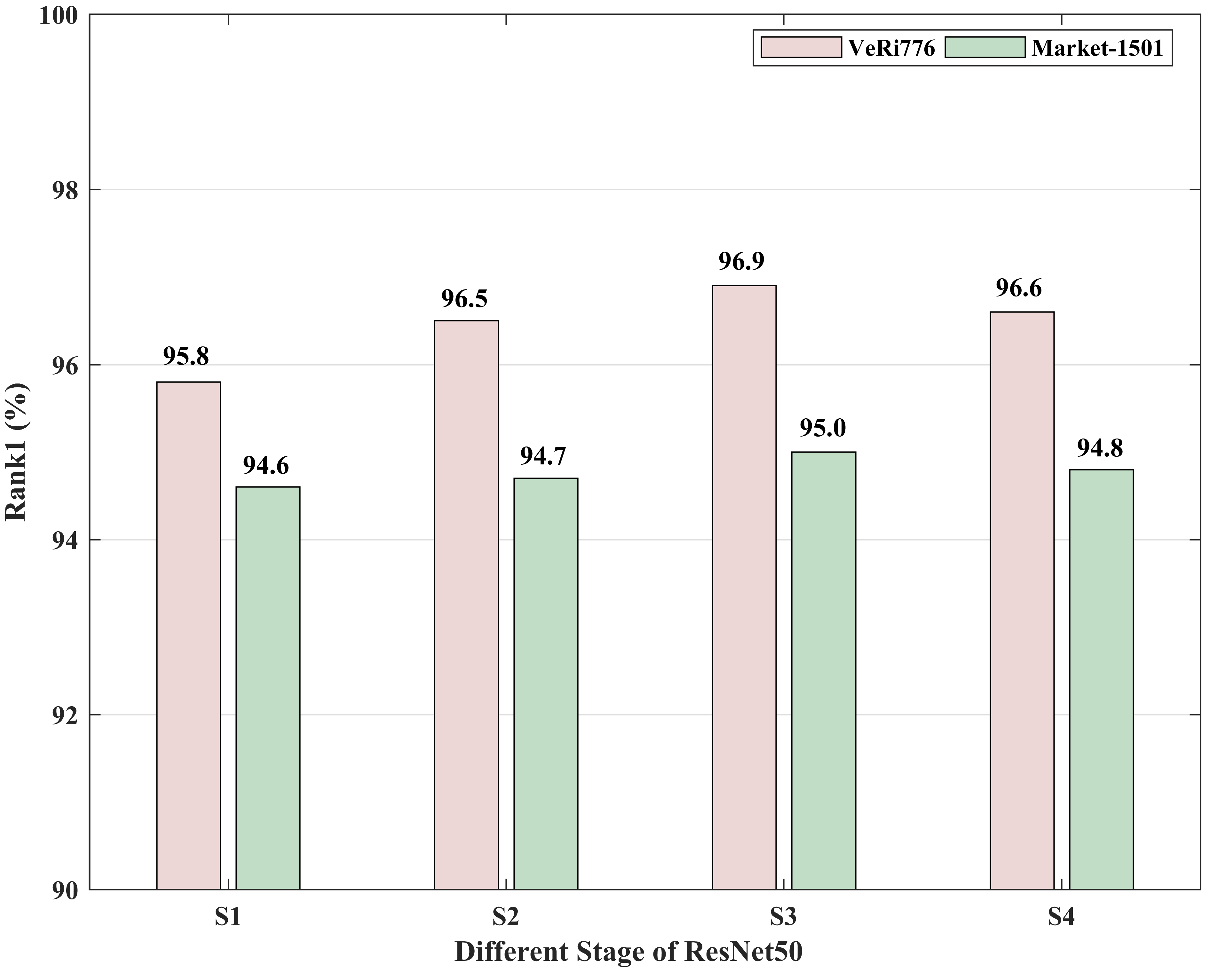}
    \caption{The Rank1 of inserting HSGM at different stage.}
    \label{fig:location_rank1}
\end{figure}

\textbf{Comparison on VehicleID}
In fact, the VehicleID \cite{drdl} dataset has a larger data scale than the VeRi776 \cite{veri776} dataset.
However, the proposed HSGNet method still can obtain the $1$st place and outperforms those state-of-the-art methods under comparison, as occurred on the VeRi776 dataset, as shown in Table \ref{tab:veri776}. For example, the proposed HSGNet method is the only case that acquires a more than 82\% rank-1 identification rate on the Test800 subset, which is respectively 3.22\% higher than that of the $2$nd places (i.e., Appearance+License \cite{app} ).
Compared with the RAM \cite{ram} of the best partition-free method, the proposed HSGNet respectively defeats it by 7.57\%, 5.51\%, and 8.95\% in terms of Rank1 identification rate on Test800, Test 1600, and Test2400 subset.
Moreover, On the largest Test 2400, compared with partition-based methods (i.e., MRM \cite{mrm} and Part Regularization \cite{partreg}), the proposed HSGNet still defeat 5.79\% and 2.45\% on Rank1 identification rate, respectively.


From Table \ref{tab:market1501} and Table \ref{tab:veri776},
 the proposed HSGNet method obtains state-of-the-art performance on Market-1501, VeRi776, and VehicleID, which shows the effectiveness and generalization of our method.
These results demonstrate that HSGNet can not only handle various scale and pose changes but also mine discriminative features.

\subsection{Ablation Studies and Analysis}
In previous subsection, we have shown the superiority of HSGNet via comparing to state-of-the-art methods.
In what’s follow, we comprehensively analyze intrinsic factors that lead to HSGNet’s superiority, including:
(1) The role of hierarchical similarity graph module (HSGM).
 (2) The effectiveness on which stage to plug the HSGM.
(3) The effect of sampling strategy of the node in HSGM.
(4) The universality for different backbones.
(5) The comparisons of parameters and computations.

\textbf{The role of hierarchical similarity graph module (HSGM).}
We conduct ablation experiments to analyze the effectiveness of different components of the hierarchical similarity graph network (HSGNet) on VeRi776 and Market-1501 datasets.
We gradually increase each component to HSGM, respectively, including the HSGM w/o Spatial and Channel Similarity Graph, HSGM w/o Hierarchical, HSGM w/o Residual (Res.) Connection, HSGM w/o Spatial Similarity Graph, HSGM w/o Channel Similarity Graph, and HSGNet.
The corresponding results are shown in Table \ref{tab:ablation}.

\begin{figure}[tp]
    \centering
    \includegraphics[width=.9\linewidth]{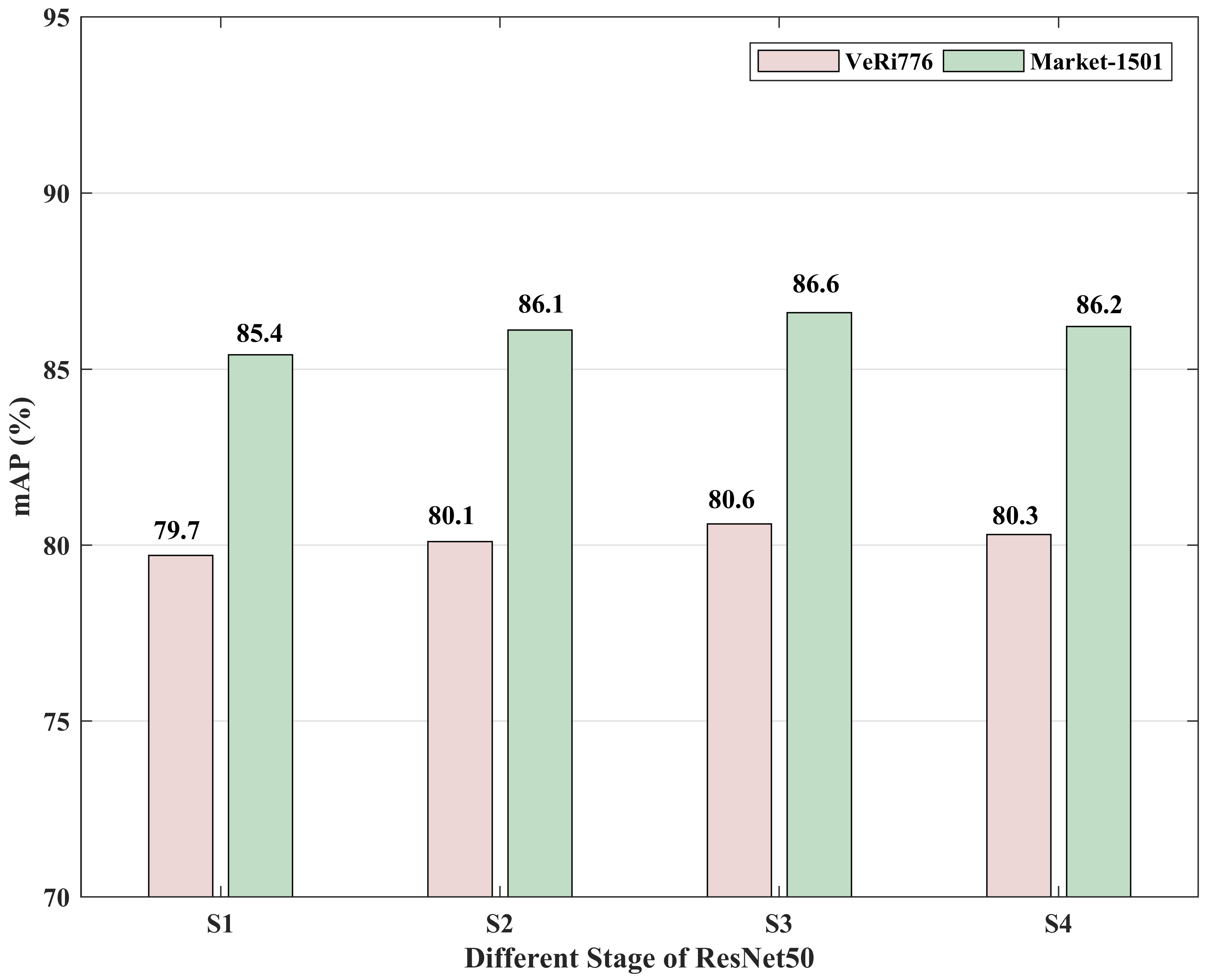}
    \caption {The mAP of inserting HSGM at different stage.}
    \label{fig:location_map}
\end{figure}

\begin{table*}[tp]
	\renewcommand{\arraystretch}{1.}
	\caption{The performance (\%) comparison among different sampling sizes of Node in HSGM.}
	\label{tab:size}
	\begin{center}
		\begin{tabular}{l|c||ccc |ccc |ccc}
			\hline
			 \multirow{2}{*}{Names}
			&\multirow{2}{*}{Sampling Sizes}
			& \multicolumn{3}{c|}{VeRi776}
			&\multicolumn{3}{c|}{Market-1501}	&\multicolumn{3}{c}{VehicleID (Average)}\\
&
			& Rank1 &Rank5     &mAP
			& Rank1 &Rank5     &mAP
			& Rank1 &Rank5     &mAP
\\
			\hline
			HSGNet1  & $k_{1}$=1, $k_{2}$=1, $k_{3}$=1
			&96.27 &98.16
			&79.60 &94.5
			&98.1 &85.6
			 &78.58
			&85.04 &81.89
  \\

			HSGNet2 & $k_{1}$=2, $k_{2}$=2, $k_{3}$=2
			&96.33 &98.27
			&79.81 &94.5
			&98.4 &85.8
 &78.90
			&85.33 &82.38
 \\
			HSGNet3 & $k_{1}$=3, $k_{2}$=3, $k_{3}$=3
			&96.58 &98.36
			&79.98 &94.7
			&98.5 &86.0
			 &79.23
			&85.68 &82.61
 \\
			HSGNet & $k_{1}$=1, $k_{2}$=2, $k_{3}$=3
			&\textbf{96.94} &\textbf{98.73}
		      &\textbf{80.65} &\textbf{95.0}
			&\textbf{98.9} &\textbf{86.6}
			 &\textbf{79.41}
			&\textbf{86.24} &\textbf{82.97}
			
 \\
			\hline
		\end{tabular}
	\end{center}
\end{table*}

\begin{table}[tp]
	\centering
	\caption{The performance of using different backbone networks.}\label{tab:backbone}
	\setlength{\tabcolsep}{2.8pt}
	\footnotesize
	\begin{tabular}{c|c||ccc|cc}
		\hline
		\multirow{2}{*}{
			\begin{tabular}{c}
			Backbone\\
			Networks\\
			\end{tabular}
		}
		& \multirow{2}{*}{
			\begin{tabular}{c}
			Using\\
			HSGM\\
			\end{tabular}
		}
		&\multicolumn{5}{c}{\multirow{1}*{VeRi776}} \\

		&&Rank1     & Rank5 &mAP & PM (M) &FLOPs (G)     \\
\hline
	\multirow{2}*{ResNet50 \cite{resnet}}
	    &  $\times$  &95.23 & 97.16  &77.31 &25.63& 4.36 \\
		&\checkmark &\textbf{96.94} & \textbf{98.73} & \textbf{80.65} &\textbf{26.85} & \textbf{4.42} \\
\hline
	\multirow{2}*{SeNet50 \cite{senet}}
	    & $\times$  &95.77 & 98.09  &79.08 &28.87& 4.41 \\
		&\checkmark &\textbf{97.16} & \textbf{98.61} & \textbf{80.67} &\textbf{30.13} & \textbf{4.47} \\
\hline
	\multirow{2}*{Res2Net50 \cite{res2net}}
	    & $\times$  &96.01 & 98.15  &78.57 &26.23&4.38 \\
		&\checkmark &\textbf{97.25} & \textbf{98.76} & \textbf{80.81} &\textbf{27.49} & \textbf{4.44} \\
\hline
	\end{tabular}
\end{table}

First, as shown in  Table \ref{tab:ablation}, it can be observed that the more settings applied in HSGM, the better performance will be acquired on both VeRi776 and Market-1501.
For example, on VeRi776 dataset, the Baseline method is lower 2.07\%, 2.72\%, 2.64\%, 2.96\%, and 3.34\% than HSGM w/o Spatial and Channel Similarity Graph, HSGM w/o Hierarchical, HSGM w/o Residual (Res.) Connection, HSGM w/o Spatial Similarity Graph, HSGM w/o Channel Similarity Graph, and HSGNet, respectively.
Second, HSGM w/o Spatial Similarity Graph and HSGM w/o Channel Similarity Graph are also better than HSGM w/o Spatial and Channel Similarity Graph in all evaluation indicators, indicating that Spatial and Channel Similarity Graph can improve and enhance the feature map for object re-identification.
Third, HSGNet also outperforms HSGM w/o Hierarchical and HSGM w/o Spatial and Channel Similarity Graph, which suggests that HSGNet can handle the difficulties of variable scales and positions.

Besides, we visualize the class activation maps of some examples from VeRi776 datasets, as shown in Figure \ref{fig:vis}. The purpose of HSGM is to mine more discriminative and recognizable features to improve the performance of object re-identification. The grad class activation maps (Grad-CAM) \cite{gradcam} visualized show that HSGNet can help the model pay more attention to local details, especially the lights, inspection signs, accessories, these are indeed differentiated information.


\textbf{The effectiveness on which stage to plug the HSGM.} We develop four methods that embed proposed HSGM at different stages of the ResNet-50's bottleneck layer.
The S1, S2, S3, and S4 respectively embed HSGM behind stage1, stage2, stage3, and stage4.
The corresponding results are shown in Fig. \ref{fig:location_rank1} and Fig. \ref{fig:location_map}.
It is observed that  no matter where the HSGM is embedded in the ResNet-50, there is a performance benefit.
For example, according to Fig. \ref{fig:location_rank1} (a), the proposed HSGM respectively brings 0.8\%, 0.9\%, 1.2\%, and 1.0\% performance improvement than Baseline method from stage1 (S1) to stage4 (S4) on Rank1 re-identification rate of the Market-1501 dataset.
As shown in  Fig. \ref{fig:location_map} (b), the proposed HSGM respectively brings 2.4\%, 2.8\%, 3.3\%, and 3.0\% performance improvement than Baseline method from stage1 (S1) to stage4 (S4) on mAP performance of the VeRi776 dataset.
Especially,  whether it is Rank1 re-identification or mAP performance, the performance improvements of HSGM in stage3 (S3) are more extensive on two datasets simultaneously.

\textbf{{The effect of sampling strategy of the node in HSGM.}}
The HSGM is a key component of HSGNet method, thus the influence of sampling strategy of Node in HSGM is evaluated.
Recalling our proposed method, the default HSGNet  adopts $k_1$=1, $k_2$=2, and $k_2$=3 sampling strategy of node.
Likewise, as shown in Table \ref{tab:size}, we design HSGMs with three different configurations, such as (1) $k_1$=1, $k_2$=1, and $k_3$=1,
(2) $k_1$=2, $k_2$=2, and $k_3$=2, (3) $k_1$=3, $k_2$=3, and $k_3$=3, namely, HSGNet1, HSGNet2, and HSGNet3, respectively.

From Table \ref{tab:size}, the HSGNet method of multi-scale sampling sizes consistently outperforms HSGNet1, HSGNet2, and HSGNet3 methods those of fixed sampling sizes on three two datasets. For example, on the VeRi776 dataset, HSGNet's mAP is 1.05\%, 0.84\% and 0.67\% higher than that of HSGNet1, HSGNet2, and HSGNet3 methods, respectively.
On the Market-1501 dataset, GiT method defeats HSGNet1, HSGNet2, and HSGNet3 methods by 1.0\%, 0.8\% and 0.6\% higher mAP performance, respectively.
These results imply that the multi-scale sampling strategy could better deal with object scale variations. Thus it is a positive strategy for HSGNet to promote object re-identification performance.

\textbf{The universality for different backbones.} As shown in Table \ref{tab:backbone}, for three well-known backbone networks (i.e., ResNet50 \cite{resnet}, SeNet50 \cite{senet} and Res2Net50 \cite{res2net}), when we embed HSGM at S3 stage, the usage of HSGM consistently increases Rank1, Rank5 and mAP performance.
For example, on the VeRi776 dataset,
the SeNet50 \cite{senet} using HSGM outperforms the SeNet50 not using HSGM by 1.39\%, 0.52\%, and 1.59\% on Rank1, Rank5, and mAP, respectively.
A similar phenomenon also occurs in Res2Net50 \cite{res2net}.  For example, when the backbone network is a Res2Net50 \cite{res2net}, the usage of HSGM obtains a 1.24\% larger Rank1 and a 2.24\% higher mAP.
These results demonstrate that HSGM has certain generality and generalization to different backbone networks, which helps to improve the performance of object re-identification.

\textbf{The comparisons of parameters and computations.}
In addition to evaluating performance, we also need to consider parameters (PM) and floating point of operations (FLOPs). From  Table \ref{tab:backbone},  the usage of HSGM brings tiny parameters (PM) and computation increments.
For example, regarding the parameter increment, the usage of HSGM additionally brings about 4.7\%, 4.3\% and 4.8\%  parameters for ResNet50 \cite{resnet}, SeNet \cite{senet} and Res2Net50 \cite{res2net}, respectively.
considering the computation increment, the usage of HSGM increases 1.37\%, 1.13\% and 1.36\% floating point of operations (FLOPs) computations for ResNet50 \cite{resnet}, SeNet \cite{senet} and Res2Net50 \cite{res2net}, respectively.
Nevertheless, comprehensive consideration of performance and computational consumption, the tiny parameter/computation increments caused by using HSGM are acceptable due to their apparent performance improvements (i.e., Rank1, Rank5, and mAP).

\section{Conclusion}\label{sec:con}

{\color{black}
This paper proposes a novel hierarchical similarity graph network (HSGNet) with a hierarchical similarity graph module (HSGM) for object re-identification.
Specifically, HSGM constructs a hierarchical graph to explore the pairwise relationships between global-local and local-local.
Then, in each hierarchical graph, the HSGM respectively regards spatial features and channel features extracted from different locations as nodes and utilizes the similarity scores between nodes to construct a spatial similarity graph and a channel similarity graph.
 Besides, we construct extensive experiments to analyze and discuss our proposed HSGNet, including:
(1) comparing a lot of state-of-the-art approaches to demonstrate the HSGNet method’s superiority;
(2) the ablation study of the HSGM;
(3) the universality of HSGM to different backbones.
(4) the comparison of parameters and computations.
Experiments on three large-scale datasets demonstrate that the proposed HSGNet outperforms state-of-the-art object re-identification approaches.	

}


%

\ifCLASSOPTIONcaptionsoff
  \newpage
\fi



%
\bibliographystyle{IEEEtran}
\small{
\bibliography{sn-bibliography}

%




\end{document}